\documentclass[twocolumn,3p]{elsarticle}
\usepackage{amsmath}
 \usepackage{amssymb}
\usepackage{lineno,hyperref}

\journal{Opt. Commun.}

\begin{document}

\begin{frontmatter}

\title{Compressive phase-only filtering at extreme compression rates}
\tnotetext[mytitlenote]{Fully documented templates are available in the elsarticle package on \href{http://www.ctan.org/tex-archive/macros/latex/contrib/elsarticle}{CTAN}.}

\author{David Pastor-Calle}
\author{Anna Pastuszczak}
\author{Micha\l{} Miko\l{}ajczyk}
\author{Rafa\l{} Koty\'{n}ski}
\cortext[mycorrespondingauthor]{Corresponding author}
\ead{rafalk@fuw.edu.pl}
\address{University of Warsaw, Faculty of Physics, Pasteura 7, Warsaw, 02-093, Poland}




\begin{abstract}
We introduce an efficient method for the reconstruction of the 
correlation between a compressively measured image and a phase-only 
filter. The proposed method is based on two properties of phase-only 
filtering: such filtering is a unitary circulant transform, and the 
correlation plane it produces is usually sparse. Thanks to these properties, phase-only filters are perfectly compatible with the framework of compressive sensing. Moreover, the lasso-based recovery algorithm is very fast when phase-only filtering is used as the compression matrix. The proposed method can be seen as a generalisation of the correlation-based pattern recognition technique, which is hereby applied directly to non-adaptively acquired compressed data. At the time of measurement, any prior knowledge of the target object for which the data will be scanned is not required. We show that images measured at extremely high compression rates may still contain sufficient information for target classification and localization, even if the compression rate is high enough, that visual recognition of the target in the reconstructed image is no longer possible. The method has been applied by us to highly undersampled measurements obtained from a single-pixel camera, with sampling based on randomly chosen Walsh-Hadamard patterns.	
\end{abstract}


\end{frontmatter}


\section{Introduction}
Linear filtering with filters that have a phase-only transfer function 
 finds numerous applications in optics ranging from  phase-only diffractive optics elements (DOE), through modeling of Fresnel diffraction, and the  spreading of a pulse envelope in a dispersive medium, up to optical pattern recognition~\cite{Goodman_Fourier_Optics}. It is rarely realized that matrices which represent discrete phase-only filtering, in contrast to matrices of other kinds of linear filtering, are always unitary. This property motivated us to further investigate the possibility of using POF as a compression matrix within the framework of compressive sensing, which is the main topic of this paper.

 Compressive sensing~\cite{IEEE_SPM_Candes2008,IEEE_SPM_24_118_Baraniuk}(CS) is a rapidly developing field of mathematics and signal processing with important contributions to the introduction of novel measurement methodologies and information recovery techniques that take advantage of the compressibility of the measured signal.
In optics, CS has been initially applied for computational ghost imaging~\cite{PRA_78_061802_Shapiro,PRA_79_053840_Bromberg,AOP_2_405_Erkmen2010,OE_20_16892_Sun}  and imaging with a single-pixel camera (SPC)~\cite{Baraniuk2008}, which allows for capturing images with a sole bucket detector rather than with a high-resolution array of detectors.
This architecture opens the way for economic electro-optical imaging systems for infrared wavelengths~\cite{Gehm2015_EO_IR}, as well as for imaging in more exotic ranges of electromagnetic radiation, such as terahertz~\cite{PIER_97_150_Yurduseven,Baraniuk_THz2008,Padilla_THz2014,APL_93_121105_Chan} or millimeter waves~\cite{Qiao2015_mm_hologr}.
 The growing range of research directions and applications based on CS 
 now include 3D imaging~\cite{OE_20_26624_Cho,Science_Sun2013}, lidar imaging~\cite{APL_101_141123_Zhao2012,Gong_2016}, joint measurement of distance to the object and its shape~\cite{Yang2013} and development of 3D laser-radar devices~\cite{PhysRevA.87.023820}.
 CS finds applications in  medical imaging~\cite{ApplOpt_54_C23_Graff} and  imaging through scattering media~\cite{SciRep_4_5552_Liutkus,Duran:15,Kolenderska:15}. It has been also applied for  spectrometry~\cite{Lan2016349}, Stokes polarimetric imaging~\cite{OL_37_824_Duran} and hyperspectral imaging~\cite{Soldevila2013,Padget_sci_rep_2015}, and holography~\cite{OL_38_2524_Clemente,AO_52_A423_Rivenson,Ramachandran2015110}. 
Image encryption with compression is another major application of CS  ~\cite{Lang201545,Zhou201510,OLT_62_152_Zhou,JMO_60_1074_Zhou,OLT_82_121_Zhou,Mehra2015344}. Ref.~\cite{IEEE_ACCESS_4_2507_Zhang} includes a recent review of CS in the field of information security.

  In this paper we focus on the possibility of using a compressive measurement of an image, in particular a measurement from a single-pixel detector, for the recovery of the spatially filtered image, when this filtered image is a lot more sparse than the original image. We show that filtering with a phase-only-filter is particularly interesting, not only because of its importance in optics but also thanks to the unitary form of the filtering matrix.
  We propose a pattern recognition method applicable to compressive measurements based on filtering with a phase-only matched filter~(POF)~\cite{Goodman_Fourier_Optics} and lasso optimization~\cite{Sampling_theory_Eldar}. This may be regarded as a refined concept of the smashed filter where a matched filter was used for target localization~\cite{ProcSPIE_6498_0H_Davenport,IEEE_TIT_59_3475_Eftekhari,RS_47_RS0N05_Volz}. In Ref.~\cite{Chen:13} it has been reported that a target object may be found at high compression rates. 
  We demonstrate that images measured compressively at extremely high compression rates, 
  at which their proper reconstruction with a quality allowing for visual examination is no longer possible, may still contain sufficient information for pattern recognition and target localization. At the same time we point to the possibility of using POF rather than other matched filters for compressive pattern recognition since it allows to obtain a very fast and efficient recovery algorithm.

\section{Mathematical background}
In this section we explain the use of linear filtering as a compression operator within the framework of compressive sensing.
\subsection{Compressive sensing}
In CS-based measurement methods, the signal or image  $\mathbf{x}$ is usually reconstructed from a compressive measurement $\mathbf{y}$
\begin{equation}
\mathbf{y}=\mathbf{M}\cdot\mathbf{x}\label{eq.measurment1},
\end{equation} 
where the size of  the signal is $n$ and the size of the measurement $\mathbf{y}$ is $m<n$. The measurement matrix $\mathbf{M}$ includes information on the measurement method and its size is $m\times n$. It is additionally assumed that the signal is compressible i.e. there exists an \textit{a priori} unknown  sparse representation of the signal
\begin{equation}
\mathbf{s}=\mathbf{T}^{-1}\cdot\mathbf{x}
\end{equation}   
with at most $k<m$ non-zero elements, whereas the size of $\mathbf{s}$ and  $\mathbf{x}$ is the same and is equal to $n$.

In order to recover the signal from the measurement it is necessary to solve 
the optimization problem such as basis pursuit denoising or lasso~\cite{Candes2006,Sampling_theory_Eldar}.
The latter is defined as 
\begin{equation}
\mathbf{\tilde s}=\operatorname*{\arg\,min}_{\mathbf{s}} \|\mathbf{A} \cdot \mathbf{s}-\mathbf{y}\|_2 \quad\textrm{subject to}\quad{ \| \mathbf{s}  \|_1 }\leq\tau \label{eq.lasso},
\end{equation}
where matrix $\mathbf{A}$ is defined as the product of the measurement and compression matrices, which should be maximally incoherent, 
\begin{equation}
\mathbf{A}=\mathbf{M}\cdot\mathbf{T}\label{eq.mt}.
\end{equation}
Then, following (\ref{eq.lasso}), $\mathbf{\tilde s}$ is determined as an approximate solution to the underdetermined set of linear equations
\begin{equation}
\mathbf{A}\cdot\mathbf{s}\approx\mathbf{y},
\end{equation}
with an imposed constraint on the $l_1$ norm $\| \mathbf{ s}  \|_1\leq\tau$. 
Lasso optimization (\ref{eq.lasso}) tends to recover a  sparse solution $\mathbf{\tilde s}$ to the  set of linear equations, or a solution which contains lots of small elements such that $\tilde s_i\approx0$.

Efficient numerical algorithms exist for solving the problem (\ref{eq.lasso}), when the matrix $\mathbf{A}$ is orthogonal or semi-orthogonal (unitary or semi-unitary for complex matrices). A rectangular and right invertible matrix is semi-unitary if its product with its complex conjugate transpose gives an identity
matrix 
\begin{equation}
\mathbf{A}\cdot\mathbf{A}^\dagger=\mathbf{I}.
\end{equation}
A straightforward way to assure the (semi)orthogonality of $\mathbf{A}$ is by selecting (semi)orthogonal matrices $\mathbf{M}$ and $\mathbf{T}$.

\subsection{Linear filtering}
We are now going to focus on matrices $\mathbf{T}$ that correspond to linear filtering. Such matrices have a Toeplitz form and in case of digital processing usually a circulant form which is a special case of a Toeplitz matrix. A $n\times n$ circulant matrix may be defined with an $n$ element vector $\mathbf{h}$ as
\begin{equation}
 T_{k,l}=n^{-1/2}\cdot{h_{((k-l)\mod{n})+1}}.
 \end{equation} 
  A circulant matrix is diagonalized by  one-dimensional discrete unitary Fourier transform (DFT) $\mathbf{F}$ (vectors and matrices in the Fourier-domain will be shortly denoted with a caret), i.e.
 \begin{equation} 
   \mathbf{\hat T}=\mathbf{F}\cdot \mathbf{T}\cdot\mathbf{F}^\dagger\label{eq.mft}
   \end{equation}   
    is a diagonal matrix with diagonal elements 
  \begin{equation}
  \hat T_{k,k}=\hat h_k\label{eq.transfer}
   \end{equation} 
   consisting of the DFT of the vector $\mathbf{h}$, i.e.
    \begin{equation}
    \mathbf{\hat h}=\mathbf{F}\cdot \mathbf{h}. 
      \end{equation} 
    Finally, a left-multiplication of a vector $\mathbf{v}$ by $\mathbf{T}$ corresponds to a circulant convolution between vectors $\mathbf{v}$ and $\mathbf{h}$ i.e. 
  \begin{equation}
  \mathbf{v}\ast\mathbf{h}=\mathbf{h}\ast\mathbf{v} =\sqrt{n}\cdot\mathbf{T}\cdot \mathbf{v}. 
  \end{equation} 
  For convenience we are using a one-dimensional notation. In two dimensions $\mathbf{F}$ is replaced by the two-dimensional DFT $\mathbf{F}\otimes\mathbf{F}$, where $\otimes$ is the Kronecker product. Images are stored in a vector form, while the circulant  matrix is now block-circulant. In a shift-invariant linear system, $\mathbf{h}$ is often called the point spread function, and $\mathbf{\hat h}$ the transfer function. 
Circulant matrices representing linear filters need not to be orthogonal (or unitary for complex matrices). However, an important class of unitary filters exists, namely the phase-only filters (POF). In fact, when the transfer function of the filter is of the form 
\begin{equation}
[\mathbf{\hat h_{POF}}]_k=\exp(i \phi_k),
 \end{equation} 
 using (\ref{eq.mft}) and (\ref{eq.transfer}) we may check that $\mathbf{T_{POF}}$ is unitary:
 \begin{equation}
 \begin{split}
\mathbf{T_{POF}}^\dagger\cdot\mathbf{T_{POF}}=
 \mathbf{F}^\dagger\cdot \mathbf{\hat T_{POF}}^\dagger\cdot\mathbf{F}\cdot\mathbf{F}^\dagger
\cdot\mathbf{\hat T_{POF}}\cdot\mathbf{F}=\mathbf{I}.
 \end{split}
 \end{equation} 
 
\subsection{Compressive phase-only filtering}
Circulant matrices defined in the Fourier space with phase-only random elements have been used for sampling in CS by Romberg~\cite{SIAM_JIS_2_1098_Romberg}. Yin \textit{et al.}~\cite{ProcSPIE_7744_0K_Yin} introduced basis pursuit and related optimization algorithms applicable to several kinds of Toeplitz and circulant matrices, however solving the optimization problem  (\ref{eq.lasso}) is simpler for unitary matrices and in consequence for POF than for other filters.
We propose to apply POF  as the compression matrix $\mathbf{T}$  in the framework of lasso optimization. Therefore equation (\ref{eq.mt}) takes the form
\begin{equation}
 \mathbf{A}=\mathbf{M}\cdot\mathbf{T_{POF}}\label{eq.a_pof}.
 \end{equation} 
Filters with a phase-only transfer function are of great importance to optics~\cite{Goodman_Fourier_Optics}. For instance, Fresnel diffraction is a kind of POF spatial filtering occurring during propagation. Indeed, the respective two-dimensional transfer function in the spatial frequency domain $(\nu_x,\nu_y)$ for the propagation at the distance $l$ equals
 \begin{equation}
\hat{h}(nu_x,nu_y)=\exp(-2 \pi i l/\lambda)\cdot \exp{\left(i\pi \lambda l(\nu_x^2+\nu_y^2)\right)}.
  \end{equation}

In  time domain the evolution of a pulse envelope is also within the second-order dispersion approximation described as POF-filtering. More general POF spatial filtering may be realized in numerous correlator-based architectures e.g.~\cite{JOSAA_31_41_Monjur}, with phase-only spatial light modulators (SLM).  Phase-only modulation makes use of the total light energy incident on an SLM, and is characterized with a high diffraction efficiency. There exist coding methods for phase diffractive optical elements (DOE) such as iterative Fourier transform algorithm  which allow to encode a rather general response in phase-only elements. Finally, POF matched filters  have been applied to optical pattern recognition constituting a recognition method with a high discrimination capacity. Phase-only filtering techniques have been successfully used for phase visualization as well as for optical encryption. 
The use of POF filtering jointly with CS in optics depends on its actual capability to produce a compressed signal representation.
Let us examine some of the aforementioned examples. For instance, a sharp image produced by Fresnel diffraction at the image plane of an imaging set-up is likely to be more sparse than the same image out of focus. A pulse envelope becomes sparse at a distance which corresponds to the highest pulse compression. In the next subsection and the rest of the paper we will focus on the POF matched filter.

\subsection{Compressive POF matched filtering}
POF filter is one of the basic filters used in optical correlation-based pattern recognition. It is defined as
\begin{equation}
[\mathbf{\hat h_{POF}}]_k={\hat r_k^*}/|\hat r_k|,
\end{equation}
where $\mathbf{r}$ is the target image.
The correlation signal with the analyzed input scene $\mathbf{x}$ is equal to 
\begin{equation}
\mathbf{s}=\mathbf{h_{POF}}*\mathbf{x}\label{eq.correl}. 
\end{equation}

The intensity distribution in $|\mathbf{s}|^2$ contains  image-recognition information with narrow detection-peaks at the locations of the recognized target images. It should be noted that the POF has been introduced to optical pattern recognition mainly because of the possibility to encode it on phase-only diffractive elements and in particular on spatial light modulators, while here we make use of its unitary form. It is also known to have a good discrimination capability but its definition does not result from a simple theoretical image formation model.

On top of that, POF-filtering is a unitary circulant transform which may be directly included in the CS framework. It is also important that  this kind of filtering tends to transform images containing a target object into a representation with a high sparsity.  

 With multiple target images stored in a dictionary $\mathbf{r}\in\{\mathbf{r}_1,\mathbf{r}_2,...\}$ it is also needed to normalize the correlation signals before they could be compared. Without further assumptions on the statistics of target occurrences,  	 $|\mathbf{s}|^2/\|\mathbf{r}_i\|^2$ may be used for target classification, where $i$ goes through the set of reference objects:
\begin{equation}
{{|\mathbf{s}_i|^2}\over {\|\mathbf{r}_i\|^2}}={{|\mathbf{h_{POF}(\mathbf{r}_i)}*\mathbf{x}|^2}\over {\|\mathbf{r}_i\|^2}}\label{eq.multi}. 
\end{equation}

Both $\mathbf{s}$ and $|\mathbf{s}|^2$ are sparse, provided that the distribution of target objects is sparse and that the cross-correlation with background objects is small.  This last example will now be further examined alongside with the pure-phase correlation (PPC) defined as
\begin{equation} 
\mathbf{s}=\mathbf{h_{POF}}*\mathbf{x'}  \quad\textrm{with}\quad {\hat x'}_k={\hat x_k}/|{\hat x_k}|,  \label{eq.white}
\end{equation}
which is also a well established pattern recognition method equivalent to POF-filtering of the whitened input scene. A compressive measurement for the PPC is based on $\mathbf{x'}$ rather than on $\mathbf{x}$, i.e.
\begin{equation}
\mathbf{y}=\mathbf{M}\cdot\mathbf{x'}\label{eq.measurment2}.
\end{equation}
  Unlike POF, PPC belongs to  nonlinear filtering techniques~\cite{Book_Sadjadi_Javidi2007} with certain adaptive properties~\cite{JOSAA_34_3915_Refregier,JOSAA_14_2162_Kotynski}. For both POF and PPC, $\mathbf{s}$ is sparse and POF is unitary, therefore it is direct to apply lasso optimization (\ref{eq.lasso}), which returns the correlation plane $\mathbf{s}$. 
If one needs to reconstruct the input scene  $\mathbf{x}$ or $\mathbf{x'}$ from the same compressive measurement, conjugate phase-only filtering may be applied to $\mathbf{s}$. However, as we will demonstrate, target recognition can be still implemented at compression rates at which visual examination of a reconstructed image is no longer possible.

\section{Numerical results}
In numerical experiments we found that lasso-based compressive matched POF filtering is particularly efficient when the compressive measurements are taken with measurement matrix $\mathbf{M}$ consisting of rows with randomly selected WH basis, while the PPC correlation gives even better results with discrete noiselet basis.  The WH transformation matrix whose size is a power of two has the following recursive form 
\begin{equation}
H_{2m}={1 \over{\sqrt{2}}} \left[ \begin{array}{cc} H_m & H_m \\ H_m & -H_m  \end{array} \right] \quad\textrm{with}\quad  H_1=1.\label{eq.WH}
\end{equation}
A respective formula for the discrete noiselet transformation is~\cite{ACHA10_27_Coifman,Pastuszczak2016}
\begin{equation}
N_{2m}={1 \over{2}} \left[ \begin{array}{cc} (1-i)\cdot N_m & (1+i)\cdot N_m \\ (1+i)\cdot N_m & (1-i)\cdot N_m  \end{array} \right],\label{eq.noiselets}
\end{equation}
with $N_1=1$.
Same as for DFT, the two-dimensional transforms are obtained through the Kronecker product of one-dimensional transforms, i.e.
\begin{equation}
H_{m\times m}^{2D}=H_m\otimes H_m,
\end{equation}
and
\begin{equation}
N_{m\times m}^{2D}=N_m\otimes N_m. 
\end{equation}

\begin{figure}[h!]
	\centering
	\includegraphics[width=0.8\linewidth]{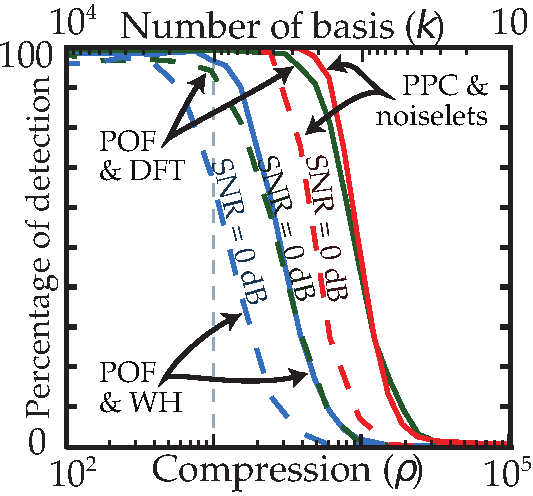}
	\caption{Compressive pattern recognition at high compression rates: Probability of correct identification and localization of the true-target obtained from a compressive measurement with  Walsh-Hadamard (WH), DFT, or noiselet sampling at various levels of the compression ratio $\rho=k/n$. Detection is based on POF-filtering for WH and DFT sampling, and on PPC for noiselet sampling.
		Probabilities shown with solid lines have been calculated for the target object in the presence of the false-target and the background, and probabilities shown with dashed lines have been obtained with an additional additive (SNR=0~dB). 
		\label{fig.recogn1}}
\end{figure}

\begin{figure}[h!]
	\centering
	\includegraphics[width=0.7\linewidth]{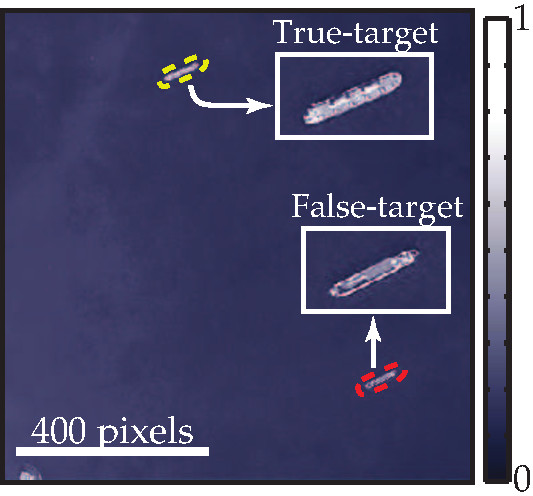}
	\caption{Synthetic 1024x1024-pixel image of the sea surface with a true-target and a false-target ship. 
		\label{fig.target}}
\end{figure}

\begin{figure}[h!]
	\centering
	(a)\includegraphics[width=0.7\linewidth]{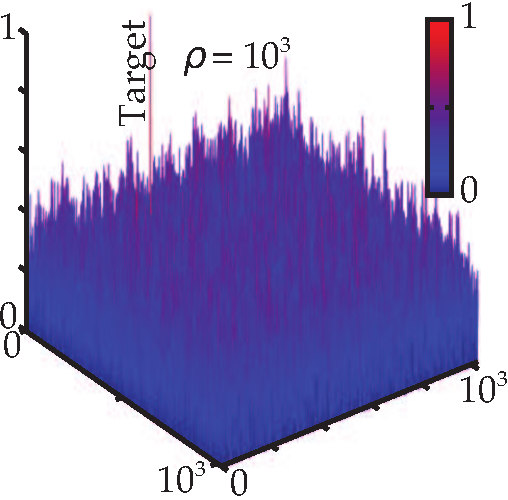}\\
	(b)\includegraphics[width=0.6\linewidth]{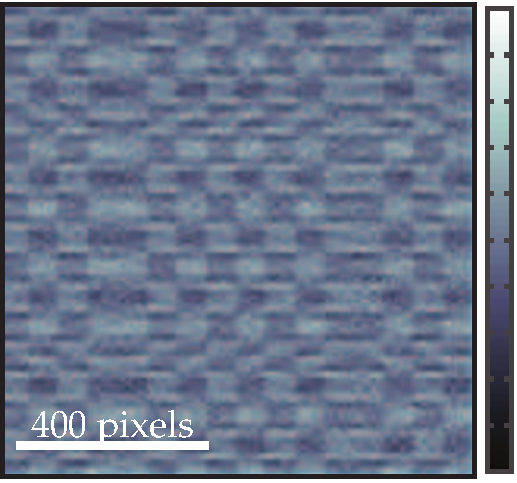}\\
	(c)\includegraphics[width=0.6\linewidth]{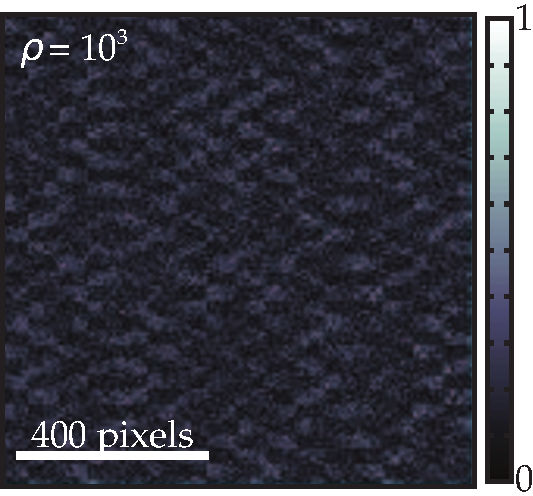}
	\caption{(a) Detection signal reconstructed with lasso and POF-filtering with WH sampling at $\rho=10^3$; (b) Indirect reconstruction of the input scene from a WH-based compressive measurement at $\rho=10^3$ obtained by conjugate POF filtering of the reconstructed correlation signal;
		(c) Direct reconstruction of the input scene from a WH-based compressive measurement at $\rho=10^3$;		
		\label{fig.recogn2}}
\end{figure}

\begin{figure}[h!]
	\centering
	(a)\includegraphics[width=0.6\linewidth]{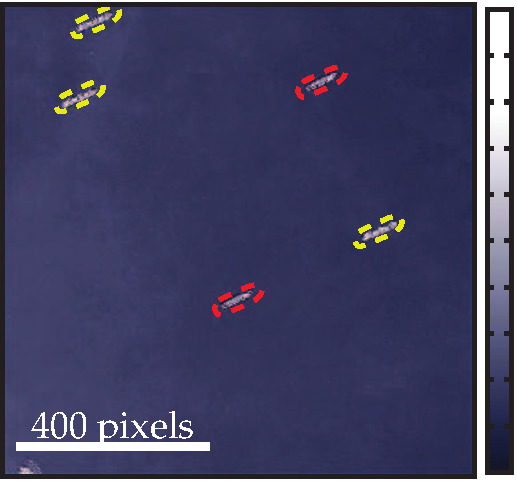}\\
	(b)\includegraphics[width=0.7\linewidth]{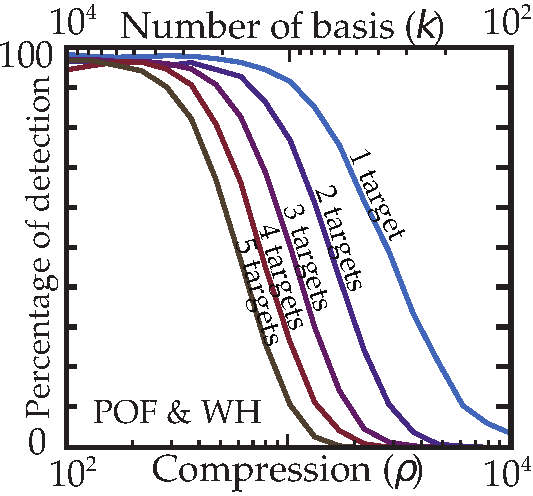}
	\caption{Compressive localization and classification in the presence of multiple targets: (a)  An example of a multi-target scene with objects selected randomly from two classes and positioned at random nonoverlapping locations; (b) Probability of correct identification and localization of all targets present in the input scene. The results have been obtained from a compressive measurement with POF and Walsh-Hadamard-based compressive measurement. 
		\label{fig.recogn3}}
\end{figure}

\begin{figure}[htbp]
	\centering
	\includegraphics[width=7.5cm]{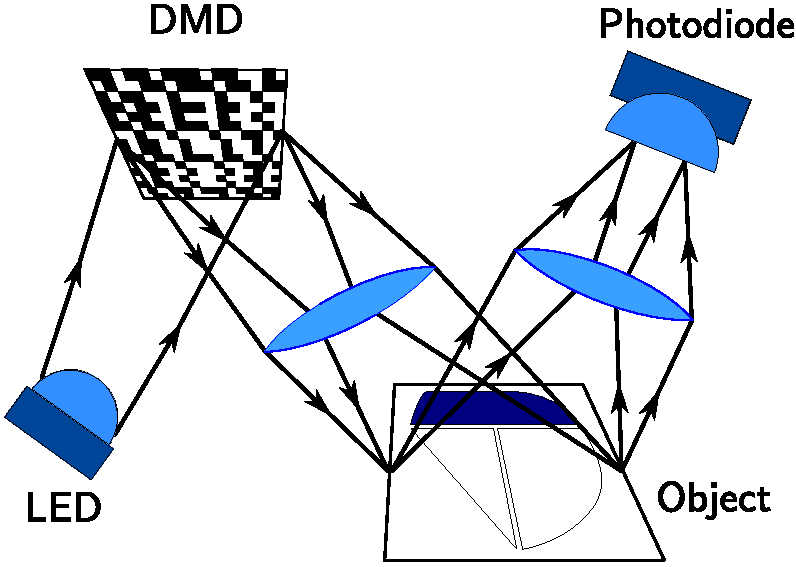}
	\caption{Schematic of the single-pixel detector.\label{fig.schem}}
\end{figure}

Small coherence between the noiselet or Walsh-Hadamard (WH) basis and the typical POF-filters makes the measurement and recognition operations independent, hence at the time of compressive measurement one does not need to have the knowledge of the target objects which the images will be later scanned for.

A measurement matrix consisting of Fourier basis could be also used with either POF or PPC, since POF-filtering has a minimal coherence with the DFT transform. 
 We include the results obtained with DFT for comparisons, however its optical implementation would be a lot more problematic. Actually, while all these methods can be easily applied numerically, only the POF-WH case can be straightforwardly implemented in an optical single-pixel detector with a binary modulator, with real non-negative images and without input scene whitening (\ref{eq.white}). 
 
 In Fig.~\ref{fig.recogn1}  we present a comparison between the three methods applied to a  $1024\times1024$ $8$-bit input scene showing the surface of sea with two ships: a target object and a false-target object. The scene is shown in Fig.~\ref{fig.target}. Both ships have a similar brightness, orientation and size.
 This scene has been compressively measured using three different methods: i)~with WH sampling - according to Eq.~(\ref{eq.measurment1}) with  matrix $\mathbf{M}$ consisting of randomly selected rows of matrix~(\ref{eq.WH}), ii)~with DFT sampling - according to Eq.~(\ref{eq.measurment1}) with  matrix $\mathbf{M}$ consisting of randomly selected DFT basis functions, and iii)~with noiselet sampling - according to Eq.~(\ref{eq.measurment2}) with matrix $\mathbf{M}$ consisting of randomly selected rows of matrix~(\ref{eq.noiselets}). Then, the reconstruction of the POF correlation signal~(\ref{eq.correl})  or PPC (\ref{eq.white}) has been obtained by solving lasso optimization~(\ref{eq.lasso}) with matrix $\mathbf{A}$ defined in Eq.~(\ref{eq.a_pof}).
 
  Each point in Fig.~\ref{fig.recogn1}  with the percentage of correct detections has been calculated based on $1000$ trials with randomly selected subsets of noiselet, WH, or Fourier basis for noise-free and noisy (additive Gaussian white noise with signal-to-noise-ratio SNR=$0$~dB) compressive measurements. 
  
  Correct recognition is possible up to extremely high compression ratios on the order of $\rho=n/k=10^3 - 10^4$.
This value is problem-dependent and varies with the size of the image and its content. In Ref.~\cite{Chen:13} for a different image, and a more general form of nonlinear correlation-based pattern recognition method, compression rates of $k/n<5\%$ (or $n/k>20$) have been reported.

We are using the SPGL1 package for lasso optimization~\cite{SIAM_JSC_vdBerg}. The algorithm has usually converged in $1$--$2$ Newton iterations with an execution time of $1$--$2$ seconds. At the same time it fails to converge for other filters than POF i.e. for non-unitary circulant operators $\mathbf{T}$. The short convergence time of the algorithm  comes in contrast to many reports of CS applications with long recovery times.  We evaluate the  noiselet and fast WH transforms in Matlab, and they are tenfold slower compared to e.g. the highly optimized fast Fourier transform (FFT) on the same computer (for a $1024\times1024$ image). Since FFT involves complex floating point arithmetic, and the numerical complexity of these transforms is similar,  therefore the convergence time could be potentially still scaled down significantly.

Theoretically, it is equivalent to recover either the original image or the POF-filtered image, as they can be easily transformed one into the other through direct or conjugate filtering. However, an approximate recovery of the (sparse) correlation plane with an  accuracy sufficient for target detection applications is possible at much higher compression rates and, in effect, it is also a lot faster. The input scene from Fig.~\ref{fig.target} could be approximately reconstructed from the WH-based compressive measurement  $\mathbf{y}$ for compression rates $\rho\lesssim 10^2$. 
However, such recovery is no longer possible at $\rho= 10^3$, while the target may still be detected. This is illustrated in Fig.~\ref{fig.recogn2}, which shows the clearly visible target detection signal reconstructed at $\rho= 10^3$ (see Fig.~\ref{fig.recogn2}(a)) alongside with the reconstructed input scene (see Fig.~\ref{fig.recogn2}(b,c), for the reconstruction from conjugate filtering, and for direct reconstruction).

 As another example we consider a combined localization and classification problem with a two-element POF dictionary and input scenes with between $1$ and $5$ randomly located non-overlapping ships of the two kinds. An example of such an input scene is shown in Fig.~\ref{fig.recogn3}(a).
 The detection is considered to be successful when all objects in the scene are correctly detected and classified. In every experiment, the detection signal is reconstructed for all reference objects from the dictionary (in our case the dictionary contains two objects) and is normalized by the respective norm of the reference object according to Eq.~(\ref{eq.multi}) to make the detection signal independent of the intensity of reference objects. The locations of the highest correlation peaks are then compared against the   actual locations of the target objects. Once again, the percentage of detection has been calculated based on $1000$ recognition experiments - for every scenario and compression ratio. 
  As may be expected, the acceptable compression rate decreases with the number of objects, as shown in Fig.~\ref{fig.recogn3}(b), nonetheless it remains high.

\section{Experimental results}
We have validated the proposed POF-based compressive target recognition method experimentally using an optical single-pixel detector shown in Fig.~\ref{fig.schem}.
We have used a $512\times 512$ pixel area of the TI LightCrafter DLP-4500 micromirror device (DMD) with a skew ($45^\circ$) orientation with respect to the display matrix to project WH patterns onto the analyzed image. This image transformation is similar as in Ref.~\cite{Lancis_dual_mode_2016} and is dictated by the way the DMD pixels are ordered and addressed within the diamond lattice.
 WH functions have been normalized to take binary values of $\{0,1\}$. Then they have been grouped into $24$-bit packets, streamed through the HDMI bus to the DLP and displayed at the frequency of $240$~Hz. Apart from the WH functions, we have also projected their negation. This allowed us to conduct a differential measurement and to eliminate an unknown bias from the experimental set-up.
The detector consisted of a photo-diode  integrated with an on-chip transimpedance amplifier (TI OPT-101P), and of a $16$-bit A/D converter (NI USB-6003 100kS/s multifunction DAQ). Its distance from the object is approximately $5$~m. The same set-up has been also described in Ref.~\cite{Pastuszczak2016} with the only difference that here we use WH bases instead of noiselets.

\begin{figure*}[h!]
	\centering
	(a)\includegraphics[width=3.2cm]{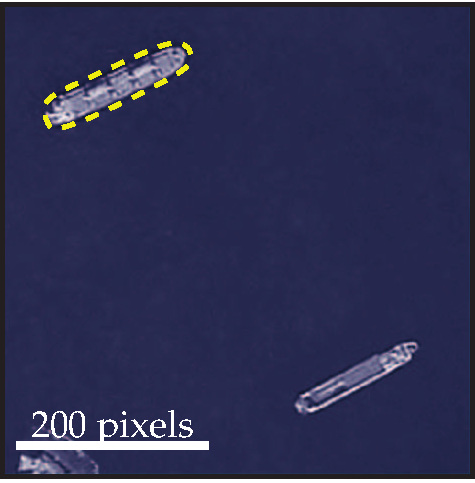}
	(b)\includegraphics[width=3.5cm]{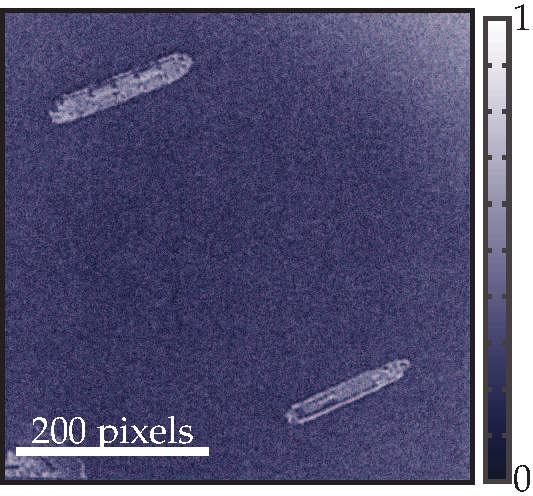}
	(c)\includegraphics[width=3.5cm]{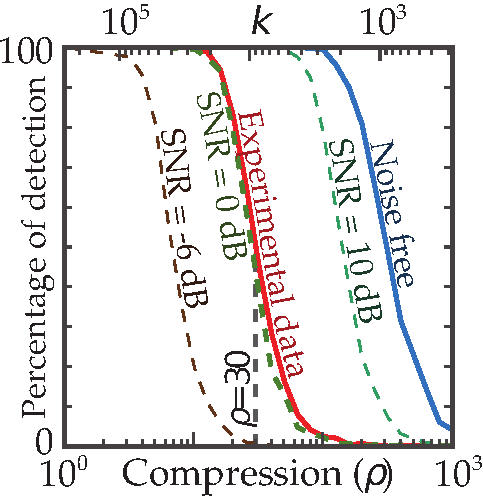}
	(d)\includegraphics[width=3.5cm]{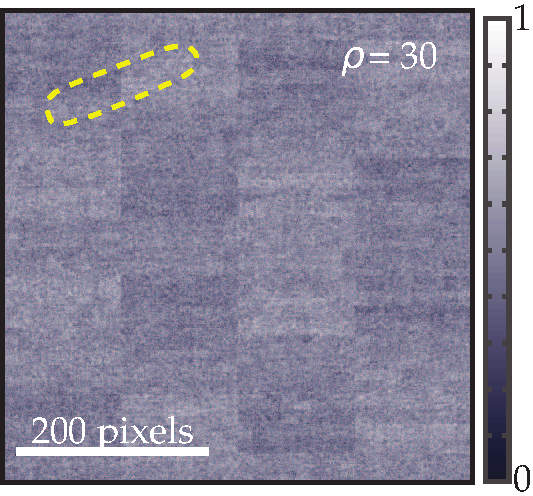}
	\caption{Experimental results from a single pixel detector: (a)~analyzed image with marked target; (b)~image reconstruction from a full $512\times 512$ WH basis; (c)~Probability of correct target detection from a compressive measurement vs compression rate obtained theoretically at various SNR levels and experimentally; (d) Signal reconstructed at $\rho=30$.\label{fig.experim}}
\end{figure*}

In Fig.~\ref{fig.experim}(a)(b) we show the $512\times 512$ synthetic image used in the experiment and its reconstruction from a full set of WH basis. 
Apart from the nonuniform illumination that could be accounted for with  simple post-processing, the low SNR on the order of $0$~dB is probably the most important limitation of the set-up. The SNR value results from the peak-to-peak amplitude of the measurements taken for different WH functions, which is approximately two orders of magnitude smaller than the average value of measured signal. This effectively reduces the useful resolution of the A/D converter by $6$-$7$ bits.
In Fig.~\ref{fig.experim}(c) we show the probability of correct target location within the radius of $5$ pixels from the exact location obtained experimentally at various compression rates. A smaller resolution than in the theoretical study presented before in  Fig.~\ref{fig.recogn1} as well as the finite SNR 
limit the acceptable level of compression. The experimental results agree with the theoretical prediction for the estimated level of SNR=$0$~dB.   Nonetheless the results clearly demonstrate that CS-based recovery of the correlation signal from an incomplete measurement is feasible. A sample reconstruction of the image at $\rho=30$ when it is still possible to detect the target but not to see it is shown in Fig~\ref{fig.experim}(d).

\section{Conclusions}
In summary, we have discussed the application of circulant matrices as compression operators for CS in optics, and we have demonstrated a compressive pattern recognition technique which allows to detect and localize a target object from a compressive measurement. It encapsulates lasso optimization, a phase-only matched filter or a pure-phase correlation adapted from optical correlation-based pattern recognition methods and a non-adaptive Walsh-Hadamard-based or noiselet-based compressive measurement. 
 We have shown in a numerical experiment that for $1024\times 1024$ pixel scenes with a relatively sparse distribution of objects in the presence of a nonuniform background it is fast and reliable at extreme  compression rates of $\rho\approx 10^3-10^4$.  The acceptable compression rate is smaller at lower resolutions, more complex images, and in the presence of noise. 
Probably the most important practical conclusion is that the application of a POF filter in a CS framework as the compression matrix leads to a very efficient recovery algorithm (we have obtained execution times on the order of $1$-$2$ seconds for $1024\times1024$-pixel images).

Like any pattern recognition method and any compressive measurement, the performance of the reconstruction methods discussed in this paper is signal-dependent. 
 The proposed method can be seen as a reimplementation of POF (or PPC) filtering for  Walsh-Hadamard-based or noiselet-based compressive measurements.  It converges to POF (PPC) at the compression rate of $1$. The study of various noise-models, or generalizations of the algorithm to other transformation invariances than shift invariance~\cite{OptExpress_15_7818_Valles}, while possible, is beyond the scope of this paper.
  
We have shown how to use the proposed method with data from an optical single-pixel detector. At the same time non-adaptive compression is also interesting for purely numerical applications, where an image database is created at some time and stored for future reference, and at the time of storage there is no prior knowledge of the target elements it will be later scanned for. 
In many areas data inflow grows faster than computational power 
and CS-based techniques, like the one discussed here, will be gaining importance. 

The proposed method can be applied to the compressive measurements in remote sensing, bioinformatics or medical imaging, in particular for high-speed screening of data from small measurement samples.


\section*{Acknowledgement}
This research was supported by Narodowe Centrum Nauki (National Science Centre, Poland UMO-2014/15/B/ST7/03107)~-~RK~\&~MM and EU 7th Framework Programme  (FP7/2007-2013, grant agreement no 316244)~-~D.P.~\&~A.P.

\end{document}